\documentclass[sn-mathphys,Numbered]{sn-jnl}

\usepackage{enumitem}
\usepackage{graphicx}%
\usepackage{multirow}%
\usepackage{amsmath,amssymb,amsfonts}%
\usepackage{amsthm}%
\usepackage{mathrsfs}%
\usepackage[title]{appendix}%
\usepackage{xcolor}%
\usepackage{textcomp}%
\usepackage{manyfoot}%
\usepackage{booktabs}%
\usepackage{algorithm}%
\usepackage{algorithmicx}%
\usepackage{algpseudocode}%
\usepackage{listings}%

\theoremstyle{thmstyleone}%
\theoremstyle{thmstyletwo}%
\usepackage{multirow}
\usepackage{enumitem}

\theoremstyle{thmstylethree}%

\begin{document}


\title[DiffSpectralNet : Unveiling the Potential of Diffusion Models for Hyperspectral Image Classification]{DiffSpectralNet : Unveiling the Potential of Diffusion Models for Hyperspectral Image Classification}


\author[1]{\fnm{Neetu} \sur{Sigger}}\email{neetu.sigger@buckingham.ac.uk}

\author*[2]{\fnm{Tuan T} \sur{Nguyen}}\email{tuan.nguyen@greenwich.ac.uk}

\author[3]{\fnm{Gianluca} \sur{Tozzi}}\email{g.tozzi@greenwich.ac.uk}
\author[4]{\fnm{Quoc-Tuan} \sur{Vien}}\email{q.vien@mdx.ac.uk}
\author[5]{\fnm{Sinh Van} \sur{Nguyen}}\email{nvsinh@hcmiu.edu.vn}

\affil*[1]{\orgdiv{School of Computing}, \orgname{University of Buckingham}, 
\orgaddress{ \country{UK}}}

\affil[2]{\orgdiv{School of Computing \& Mathematical Sciences}, \orgname{University of Greenwich}, 
\orgaddress{\country{UK}}}
\affil[3]{\orgdiv{School of Engineering}, \orgname{University of Greenwich}, \orgaddress{ \country{UK}}}
\affil[4]{\orgdiv{Faculty of Science and Technology}, \orgname{Middlesex University}, \orgaddress{ \country{UK}}}
\affil[5]{\orgdiv{School of Computer Science and Engineering}, \orgname{International University–Vietnam National University of HCMC}, \orgaddress{\country{VN}}}


\abstract{

Hyperspectral images (HSI) have become popular for analysing remotely sensed images in multiple domain like agriculture, medical. However, existing models struggle with complex relationships and characteristics of spectral-spatial data due to the multi-band nature and data redundancy of hyperspectral data. To address this limitation, we propose a new network called DiffSpectralNet, which combines diffusion and transformer techniques. Our approach involves a two-step process. First, we use an unsupervised learning framework based on the diffusion model to extract both high-level and low-level spectral-spatial features. The diffusion method is capable of extracting diverse and meaningful spectral-spatial features, leading to improvement in HSI classification. Then, we employ a pretrained denoising U-Net to extract intermediate hierarchical features for classification. Finally, we use a supervised transformer-based classifier to perform the HSI classification. Through comprehensive experiments on HSI datasets, we evaluate the classification performance of DiffSpectralNet. The results demonstrate that our framework significantly outperforms existing approaches, achieving state-of-the-art performance.
}




\maketitle

\section{Introduction}\label{sec1}

Hyperspectral Images (HSI) are now being captured more effectively by imaging spectrometers aboard satellites and aircraft. Unlike regular optical images with just three channels (e.g., Red, Green, Blue), each pixel of HSI contains abundant and continuous spectral information. This allows for the identification of intricate spectral characteristics of subjects that might otherwise go unnoticed. HSI is extensively used in various Earth remote sensing applications, including land use and land cover classification \cite{article}, precision agriculture \cite{rs12162659}, object detection \cite{8738045}, tree species classification, brain cancer detection \cite{10.1371/journal.pone.0193721}, and more.


The challenges of classification in HSI arise from their high dimensionality, strong correlations between adjacent bands, a nonlinear data structure, and limited training samples \cite{Paoletti2019DeepLC}. To address these challenges and improve classification accuracy, researchers have proposed several methods. While traditional approaches like Maximum Likelihood Classification have been foundational, they often face challenges with high-dimensional data spaces, known as the curse of dimensionality \cite{Hughes1968OnTM}.

Initially, spectral information for each pixel was fed into neural networks to identify the corresponding class \cite{572944}. As data dimensionality increased, feature selection and dimensionality reduction became crucial. Techniques like Principal Component Analysis (PCA) \cite{Rodarmel2002PrincipalCA} and SVM \cite{6297992} were often employed to achieve better classification results. However, traditional neural networks faced difficulties in effectively utilising the spatial-spectral relationships and capturing complex information in HSI.

Following this development, Convolutional Neural Networks (CNNs) have been shown to be more effective in HSI classification compared to neural networks due to their ability to integrate spectral and spatial-contextual information in the classification process. CNNs have better feature representation and high accuracy in classification and have demonstrated promising performance in HSI classification. CNNs can automatically extract hierarchical features from HSIs \cite{zeng2020semisupervised}. As datasets grew, deeper architectures like Residual Networks (ResNets) were introduced, specifically adapted to capture complex patterns in HSI data for classification \cite{rs14132997}. Advanced architectures such as autoencoders were later developed to extract a compressed representation of HSI data for classification purposes \cite{6782778}. Attention mechanisms were integrated into CNN architectures to enhance the accuracy of classification by weighing the importance of different spectral bands \cite{hang2020hyperspectral}. Furthermore, advancements in CNNs led to the introduction of novel pooling and unpooling mechanisms that better preserve spatial information during classification \cite{rs13050930}.

While RNNs \cite{Mou2017DeepRN} are capable of capturing the spatial-spectral relationship from long-range sequence data, they face challenges such as vanishing gradients and dependency on the order of spectral bands. Transformers, originally designed for natural language processing (NLP), have shown promising results when integrated into HSI classification. They effectively capture long-range dependencies in hyperspectral data \cite{Hong_2022, rs15153721}. To overcome the limitations of CNNs in pixelwise remote sensing classification and spectral sequence representation \cite{ijerph20043059}, a multispectral image classification framework was introduced. This framework integrates Fully Connected (FC) layers, CNNs, and Transformers. Unlike the classic transformers that focus on band-wise representations, SpectralFormer \cite{Hong_2022} is an example of such a framework that captures spectrally local sequence information, creates group-wise spectral embeddings, and introduces cross-layer skip connections to retain crucial information across layers through adaptive residual fusion. Another novel model, SS1DSwin \cite{10178075}, is based on Transformers and implements the network architecture of Swin Transformer. It effectively captures reliable spatial and spectral dependencies for HSI classification.

Deep Neural Networks (DNNs) show promise in HSI classification but struggle to effectively model spectral-spatial relationships. Transformer-based methods generally do not fully leverage spatial information \cite{Sun2022SpectralSpatialFT} and have limitations in extracting fine-grained local feature patterns \cite{gulati2020conformer}. Effectively learning rich representations and addressing the complexities of spectral-spatial relations in high-dimensional data are crucial for achieving optimal HSI classification results. In conclusion, Transformer-based methods face challenges in directly capturing reliable and informative spatial-spectral representations available in HSI for efficient and robust classification.


To address these issues, we have reevaluated the process of extracting features from the HSI data from different perspectives. As a result, we have developed a new HSI classification method that incorporates diffusion and transformer techniques leveraging their respective advantages. The features representation learned from the diffusion models have been demonstrated to be highly effective in various discriminating tasks with impressive performance like semantic segmentation \cite{baranchuk2022labelefficient}, object detection \cite{Chen2023DiffusionMF}, and face generation \cite{perera2023analyzing}.


This paper presents a novel classification framework called the diffusion-based spectral-spatial network combined with transformers. Its aim is to capture the spectral-spatial relationship from HSI, obtain deep features that are both effective and efficient, and fully utilize the spectral-spatial information of the data. The main contributions of this paper are summarized as follows:
\begin{enumerate} [label=\arabic*)]
 \item By integrating forward and reverse diffusion processes, our proposed framework exploits the diffusion model to extract unsupervised spectral-spatial features from HSI data. The utilization of both processes facilitates the acquisition of high-level and low-level features
\item To effectively and efficiently utilize the abundant timestep-wise features, we extract intermediate hierarchical features from the denoising U-Net at different timesteps. Subsequently, we employ a proposed supervised transformer-based classifier for performing HSI classification.
\item Experiments conducted on three widely known datasets demonstrate that the proposed DiffSpectralNet method yields significant improvements in classification results and outoerforms other advanced HSI classification methods. in term of overall accuracy, average accuracy, and Kappa coefficient.
\end{enumerate}

\section{Research Methodology}\label{sec2}
We have developed a method called DiffSpectralNet that consists of two stages: an unsupervised diffusion process and a supervised classification.
The unsupervised diffusion process is based on the denoising diffusion probabilistic model (DDPM) \cite{DBLP:journals/corr/abs-2006-11239} for the purpose is to learn spectral-spatial representations effectively. In this process, we extract plenty of spectral-spatial features from various time steps \textit{t} during the reverse diffusion process of DDPM to capture the characteristics of different objects in HSI data. Finally, these features are inputted into the supervised classification model for classification.


\subsection{Diffusion-based Unsupervised Spectral-Spatial Feature Learning}
In order to capture complex spectral-spatial relations and label-agnostic information of HSI data effectively, the first step of our proposed approach is to train a diffusion model in an unsupervised manner. Then, we introduce the detailed formulation of our unsupervised feature learning procedure, which involves diffusion-based forward and backward processes with the HSI data.

\begin{figure}[t]
    \centering
    \includegraphics[clip, trim=0cm 0cm 0cm 0cm, width=0.99\textwidth]{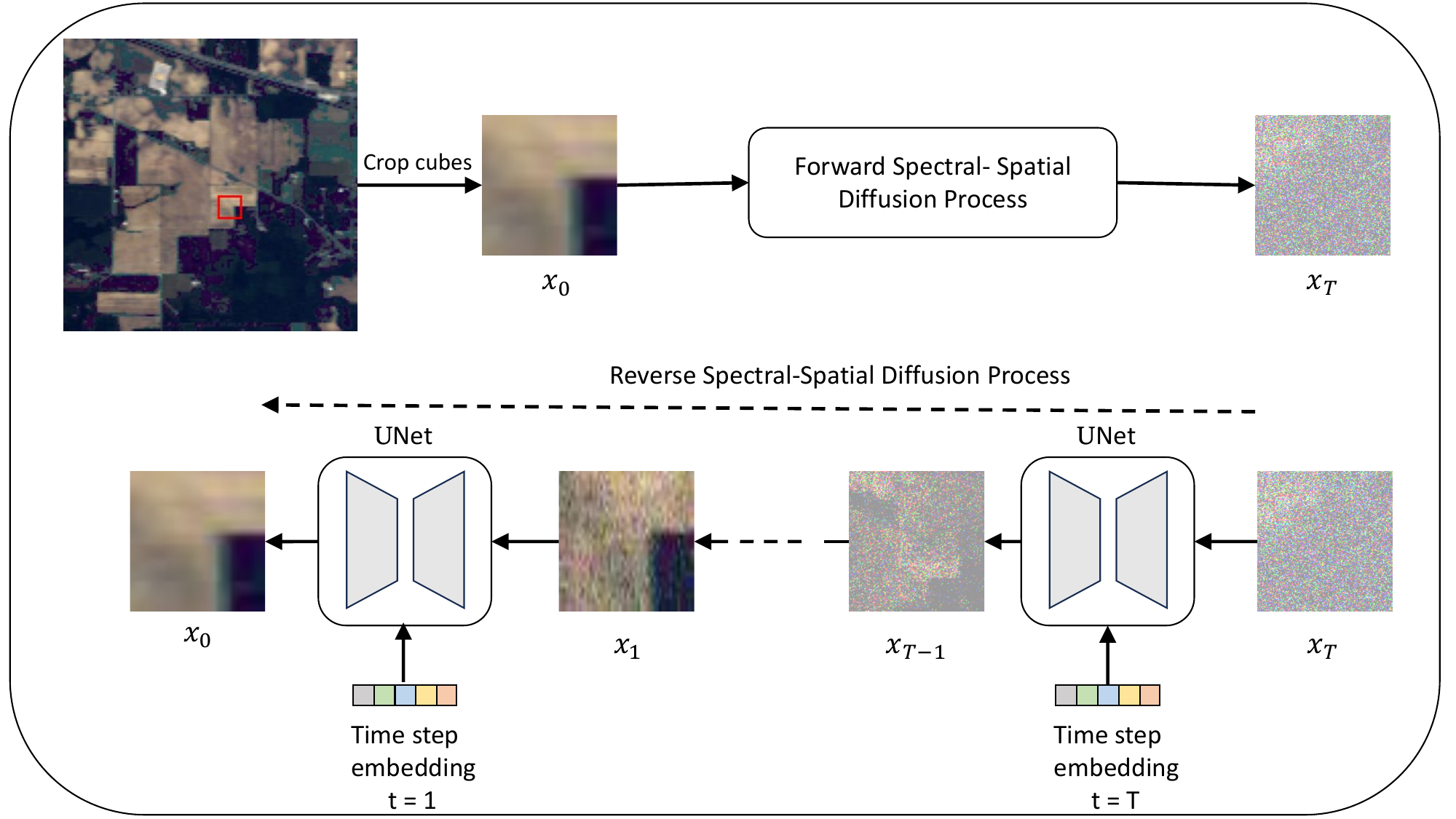}
    \caption{Overview of our proposed Unsupervised Spectral-Spatial Feature
Learning Network, Unsupervised Spectral-Spatial Feature Learning.}
    \label{fig:diffusion}
\end{figure}

\begin{enumerate} [label=\arabic*)]
\item	\textit{Forward Diffusion Process:} 
DDPM represents a category of models based on likelihood estimations. In the forward process, Gaussian noise is added to the original training data. In our proposed model, we aim to learn spectral-spatial features effectively in an unsupervised manner. We start by training our DDPM  using unlabeled patches randomly cropped from the HSI dataset. To prepare the data for training, the data is pre-processed by patch cropping operation. Next, patches are randomly sampled from HSI for DDPM training. Formally, given an unlabeled patch \(x_0 \in \mathbb{R}^{P \times P \times B}\), where \textit{P} denote the height and width of patch \(x_0 \), \textit{B} represents the number of spectral channels, respectively. During the forward diffusion process, Gaussian noise is gradually added to the HSI patch according to the variance schedule \(\{\beta_t\}_{t=0}^{T}\) in the diffusion process where \textit{T} is the total number of the timestep. The process follows the  Markov chain \cite{DBLP:journals/corr/abs-2006-11239} process:

\begin{equation}
q(x_t|x_{t-1}) = \mathcal{N}\left(\sqrt{(1 - \beta_t)}x_{t-1}, {\beta_t}\textit{I}\right)
  \tag{1}
\end{equation}
where $\mathcal{N}$ is a Gaussian distribution. The above formulation leads to the probability distribution of the HSI at a given time $\textit{t} + 1$ is obtained by its state at time \textit{t}. During the first diffusion, the spectral-spatial instance with noise is expressed as follows:
\begin{equation}
x_{1} = \sqrt{\alpha_{1}}x_{0} + \sqrt{1 - \alpha_{1}}\varepsilon
  \tag{2}
\end{equation}

At the \(\textit{t}_{th}\) step, the spectral-spatial instance incorporated with noise is expressed as follows: 
\begin{equation}
   x_t = \sqrt{\overline{\alpha}_t} x_0 + \sqrt{1 - \overline{\alpha}_t} \epsilon, \quad\epsilon\sim \mathcal{N}(0, \textit{I})
     \tag{3}
     \label{eq3}
\end{equation}
where \(\alpha_t\) = \(1 - \beta_t\) and, \(\overline{\alpha}_t\) represents the product of \(\alpha_1\) to \(\alpha_t\). Given these inputs, the hyperspectral instance at timestep \textit{t} can be straightforwardly produced by equation \ref{eq3}.\\

\item \textit{Reverse Diffusion Process:} In the reverse diffusion process, a spectral-spatial \textbf{U}-Net \cite{saharia2021image} denoising network is employed as shown in \textcolor{red}{Fig.  }is trained to predict the noise added on $x_{t-1}$, taking noisy patch $x_t$ and timestep $t$ as inputs. And $x_{t-1}$ is calculated by subtracting the predicted noise from $x_t$. DDPM uses a Markov chain process to remove the noisy sample $x_T$ to $x_0$ step by step. Under large $T$ and small $\beta_t$, the probability of reverse transitions is approximated as a Gaussian distribution and is predicted by a \textbf{U}-Net as follows:
\begin{equation}  
p_{\theta}(x_{t-1}|x_{t}) = \mathcal{N} \left( x_{t-1}; \mu_{\theta}(x_{t}, t), \sigma_{\theta}(x_{t}, t) \right) \tag{4}
\end{equation}

where the reverse process can be re-parameterized by estimating $\mu_{\theta}(x_{t}, t)$ and $\sigma_{\theta}(x_{t}, t)$. $\sigma_{\theta}(x_{t}, t)$ is set to $\sigma_t^2 I$, where $\sigma_t^2$ is not learned. To obtain the mean of the conditional distribution \(p_{\theta}(x_{t-1}|x_{t})\), we need to train the network to predict the added noise.The mean of $\mu_{\theta}(x_{t}, t)$ is derived as follows:
\begin{equation}
    \mu_{\theta}(x_{t}, t) = \frac{1}{\sqrt{\alpha_t}} \left( x_{t} - \frac{1 - \alpha_t}{\sqrt{1 - \alpha_t}} \epsilon_{\theta}(x_{t}, t) \right)
    \tag{5}
\end{equation}
where $\epsilon_{\theta}(\cdot, \cdot)$ denote the spectral-spatial denoising network whose input is the timestep $t$ and the noisy hyperspectral instance $x_t$ at timestep $t$. \\
The denoising network takes in the noisy hyperspectral instance along with the timestep to produce the predicted noise. The \textbf{U}-Net denoising model $\epsilon_{\theta}(x_{t}, t)$ is optimized by minimizing the loss function of the spectral-spatial diffusion process can be expressed as follows:
\begin{equation}
\mathcal{L}(\theta) = \mathbb{E}_{t, x_0, \epsilon} \left[ \left( \epsilon - \epsilon_{\theta} \left( \sqrt{\overline{\alpha_t}} x_0 + \sqrt{1 - \overline{\alpha_t}} \epsilon, t \right) \right)^2 \right]
\tag{6}
\end{equation}
\end{enumerate}
\subsection{Supervised Classification using Spectral-Spatial Diffusion Feature}
After training the network using Unsupervised Spectral-Spatial methods, we start extracting useful diffusion features from the pre-trained DDPM. Next, we employ a transformer-based classifier for classification.

During the feature extraction step, we utilize the \textbf{U}-Net denoising network to extract a spectral-spatial timestep-wise feature. The pre-training of DDPM enables it to capture rich and divers information from the input data during the reverse process. As a result, we extract features from the intermediate hierarchies of DDPM at various timesteps to create robust representations that encapsulate the salient features of the input HSI.


The parameters of the pre-trained DDPM remain constant, as shown in Fig... We gradually add Gaussian noise to the input patch \( x_0 \in\mathbb{R}^{P \times P }\) through the diffusion process. For a noisy input patch \( x_t \) at timestep \( t \), the noisy version \( x_t \) can be directly determined using equation \ref{eq3}. Subsequently, \( x_t \) is fed into the pre-trained spectral-spatial denoising \textbf{U}-Net to derive hierarchical features from the \textbf{U}-Net decoder. Diffusion features from various decoder layers are collectively upsampled to \( P \times P \) and then merged to form the feature \( f_t \) in \(\mathbb{R}^{P \times P \times L}\) at timestep \( t \), where \( P \) represents the height and width of the patch and \( L \) denotes the feature channel. For each feature \( f_{ti}  \in \mathbb{R}^{P \times P \times L}\), we retain only the vector associated with the center pixel, indexed as \( C_i  \in \mathbb{R}^{p \times p \times L}\). This approach significantly reduces the computational cost due to a decrease in parameters.

We construct the \( n \)-set of timestep-wise spectral-spatial feature repository using the extracted diffusion feature, which is represented as \(\beta =  C(f_{ti}) \), where \( i \) is \{1, \ldots, n\}. The timesteps \( t_1, \ldots, t_m \) are sampled from the interval $[0, T]$ at equal intervals. The extracted features \(C(f_{ti}) \) use a linear projection layer for mapping features to a token sequence for the transformer. Positional embedding is added to the input token sequence before feeding it to the transformer. This provides the transformer with information about the relative positions of the patches. Therefore, the abundant features contain diverse and multi-level information of the input HSI data, which we use for classification.


A network is needed to predict the classification label after mapping the patch representation. Transformer-based classifiers are trained based on the inspiration from \cite{Hong_2022}, as shown in the Fig... These classifiers take positionally embedded image patches as inputs and use an MLP head to predict the final classification scores. The skip connection-based classification module combines the CNN and transformer structures to form an effective classifier. This approach utilizes skip connection to enhance the information transitivity between layers, multi-head attention mechanisms, feed-forward neural networks to spectral-spatial feature mapping, and a Transformer structure for deep feature extraction, resulting in outstanding classification performance.

\section{Experimental Results and Analysis}\label{sec3}
In this section, we begin by introducing three experimental HSI datasets. Following that, we describe the experimental setting, which includes evaluation metrics and implementation details. Finally, we compare the results with some state-of-the-art deep learning methods, clearly proving the advantages of the proposed algorithm.

\subsection{Hyperspectral Datasets Description}\label{subsec2} 
\begin{enumerate} [label=\arabic*)]
    \item \textit{Indian Pines (IP): } The IP dataset was collected in 1992 using the Airborne Visible Infrared Imaging Spectrometer (AVIRIS) Sensor, covering the northwestern region of Indiana in the United States. It consists of $145×145$ pixels with a spatial resolution of $20 m$ and $220$ spectral bands in the wavelength range of $400 to 2500 nm$. The dataset contains labeled pixels with 16 categories. The class name and the number of training and testing samples are listed in Table \ref{table1_indian}.
\begin{table}[h]
\caption{Land Cover Classes with the Standard Training and Testing Samples for the Indian Pines Dataset}\label{table1_indian}
\begin{tabular}{clcc}
\toprule
Class No. & Class Name & Training & Testing \\
\midrule
1 & Alfalfa & 5 & 41 \\
2 & Corn-notill & 143 & 1285 \\
3 & Corn-mintill & 83 & 747 \\
4 & Corn & 24 & 213 \\
5 & Grass-pasture & 48 & 435 \\
6 & Grass-trees & 73 & 657 \\
7 & Grass-pasture-mowed & 3 & 25 \\
8 & Hay-windrowed & 48 & 430 \\
9 & Oats & 2 & 18 \\
10 & Soybean-notill & 97 & 875 \\
11 & Soybean-mintill & 245 & 2210 \\
12 & Soybean-clean & 59 & 534 \\
13 & Wheat & 20 & 185 \\
14 & Woods & 126 & 1139 \\
15 & Buildings-Grass-Trees-Drives & 39 & 347 \\
16 & Stone-Steel-Towers & 9 & 84 \\
\midrule
& \textbf{Total} & \textbf{924} & \textbf{8215} \\
\botrule
\end{tabular}
\end{table}

 \item \textit{Pavia University (PU): }The second HSI dataset is the well-known PU, acquired by the Reflective Optics System Imaging Spectrometer (ROSIS) sensor. The ROSIS sensor acquired $103$ bands covering the spectral range from $430 to 860 nm$, and the dataset consists of $610 × 340$ pixels at GSD of $1.3 m$. Moreover, there are 9 land cover classes in the dataset. The class name and the number of training and test sets are detailed in Table \ref{table2_pavia}.
\begin{table}[h]
    \centering
    \caption{Land Cover Classes with the Standard Training and Testing samples for the Pavia University Dataset}
    \label{table2_pavia}
    \begin{tabular}{cccc}
        \toprule
        Class No. & Class Name            & Training & Testing \\
        \midrule
        1        & Asphalt               & 332      & 6299   \\
        2        & Meadows               & 932      & 17717  \\
        3        & Gravel                & 105      & 1994   \\
        4        & Trees                 & 153      & 2911   \\
        5        & Painted metal sheets  & 67       & 1278   \\
        6        & Bare Soil             & 251      & 4778   \\
        7        & Bitumen               & 67       & 1263   \\
        8        & Self-Blocking Bricks  & 184      & 3498   \\
        9        & Shadows               & 47       & 900    \\
        \midrule
        Total    &                       & 2037     & 36558  \\
        \bottomrule
    \end{tabular}
\end{table}

\item \textit{Salinas Scene (SS): } The SS dataset was collected using the AVIRIS sensor and is situated in Salinas Valley, California. The spatial resolution is set at $3.7 m$. and the dataset includes 16 crop types and has been widely utilized in classification. After the exclusion of 20 bands associated with water vapor and noise, a total of 204 bands remained, resulting in a data size of \(512 \times 217\). The detailed breakdown of land cover types and their respective pixel counts is presented in Table \ref{table3_salinas}.
 
\begin{table}[h]
    \centering
    \caption{Land Cover Classes with Training and Testing samples for the Salinas scene Dataset}
    \label{table3_salinas}
    \begin{tabular}{cccc}
        \hline
        Class No. & Class Name                         & Training & Testing \\
        \hline
        1        & Brocoli\_green\_weeds\_1            & 100      & 1909   \\
        2        & Brocoli\_green\_weeds\_2            & 186      & 3540   \\
        3        & Fallow                             & 98       & 1878   \\
        4        & Fallow\_rough\_plow                & 69       & 1325   \\
        5        & Fallow\_smooth                     & 133      & 2545   \\
        6        & Stubble                            & 197      & 3762   \\
        7        & Celery                             & 178      & 3401   \\
        8        & Grapes\_untrained                  & 563      & 10708  \\
        9        & Soil\_vinyard\_develop             & 310      & 5893   \\
        10       & Corn\_senesced\_green\_weeds       & 163      & 3115   \\
        11       & Lettuce\_romaine\_4wk              & 53       & 1015   \\
        12       & Lettuce\_romaine\_5wk              & 96       & 1831   \\
        13       & Lettuce\_romaine\_6wk              & 45       & 871    \\
        14       & Lettuce\_romaine\_7wk              & 53       & 1017   \\
        15       & Vinyard\_untrained                 & 363      & 6905   \\
        16       & Vinyard\_vertical\_trellis         & 90       & 1717   \\
        \hline
        Total    &                                   & 2548     & 65592  \\
        \hline
    \end{tabular}
\end{table}
\end{enumerate}

\subsection{ Parameter Setting and Analysis}\label{subsubsec2}
\begin{enumerate}[label=\arabic*)]
\item \textit{Evaluation Metrics: } We evaluate the performance using three prominent metrics: overall accuracy (OA), average accuracy (AA), and Kappa coefficient ($\kappa$). OA gives a direct insight into general model performance, and AA ensures each class has a balanced contribution, especially in imbalanced datasets, while $\kappa$ measures the reliability between the ground truth and model predictions.
\item \textit{Implementation Details: } We used the PyTorch framework to implement and train the DiffTrans-HSI model. The training was done on a basic hardware setup, which consists of a POWER8NVL production-grade CPU with 128 CPU threads spread across 2 sockets for efficient processing. Additionally, four NVIDIA Tesla P100 GPUs were used for enhanced graphical computations, each offering a memory of approximately 16 GB.
The pre-training procedure for the diffusion model was optimized using the Adam optimizer. We set the learning rate to $1 \times 10^{-4}$, with a batch size of 128 and a patch size of \(32 \times 32\). We trained for 30,000 epochs for all datasets. In the second stage, we trained the classification model using the Adam optimizer, maintaining the same learning rate of $1 \times 10^{-4}$ and a batch size of 128. Due to hardware limitations, we use batch size 64 for the SS dataset. To determine the amount of spectral information preserved in the compressed data, we employed PCA. Given that, each dataset presents a distinct number of features post pre-training with the diffusion model, the range of PCA components varies among the three datasets. The number of epochs was set to 300 for IP and 600 for PU and SS datasets.
\item \textit{Quantitative Results and Analysis: } To demonstrate the effectiveness of our proposed DiffSpectralNet, we compare our classification performance with various state-of-the-art approaches, and the following methods were chosen: FuNet-C \cite{Hong2020GraphCN}, DMVL \cite{9254125}, 3DCAE \cite{8694830}, RPNet–RF \cite{s23052499}, GSSCRC \cite{electronics12183777}, JPPAL\_CRF \cite{rs15163936} and SS1DSwin \cite{10178075}. Note that, we directly use the results of each of these methods as reported in their papers. All of these methods, CNN-based and transformer-based methods, produce good classification results. A detailed overview for the compared methods is presented below.

\begin{itemize}
    \item The FuNet-C \cite{Hong2020GraphCN} proposed minibatch GCN (miniGCN), designed to train large-scale GCNs in a minibatch fashion. miniGCN is one of the classical algorithms for constructing sample relationships.  
    \item The DMVL \cite{9254125}, similar to our proposed model, follows the two-stage algorithms. It performs unsupervised feature extraction followed by classification using an SVM classifier. 
    \item The 3DCAE \cite{8694830}  is an unsupervised method to learn spectral-spatial features. It uses the encoder-decoder backbone with 3D convolution operations.
    \item The RPNet–RF \cite{s23052499} merges Random Patches Network (RPNet) with Recursive Filtering (RF) for deep feature extraction. Initially, image bands undergo convolution with random patches for multi-level RPNet features, which are refined using RF after dimension reduction via PCA. These enhanced RPNet–RF features and HSI spectral features facilitate HSI classification through an SVM classifier.
    \item GSSCRC  \cite{electronics12183777} algorithm incorporates the cooperative representation classification model and introduces the geodesic distance calculation method to select spectral nearest-neighbor information, thereby effectively utilizing the neighbor information in hyperspectral images. This approach facilitates the exploration and utilization of the spatial–spectral neighborhood structure of hyperspectral data for HSI classification.
    \item JPPAL\_CRF \cite{rs15163936} utilizing the entire posterior probability matrix for enhanced sample selection. This method enhances sample variability, and optimise conditional random fields (CRF) results by harnessing spatial data and label constraints.
    \item SS1DSwin \cite{10178075} design reveals local and hierarchical spatial–spectral links through two modules: the Groupwise Feature Tokenization Module (GFTM) and the 1DSwin Transformer with Cross-Block Normalized Connection Module (TCNCM). GFTM processes overlapping cubes and uses multihead self-attention for spatial–spectral relationships. Meanwhile, TCNCM utilizes window-based strategies for spectral relationships and cross-block feature fusion.\\
\end{itemize}
\end{enumerate} 

Based on the analysis of classification results obtained for the IP, PU, and SS datasets presented in Table \ref{IP_results}, Table \ref{pavia_results}, and Table \ref{sainas_results}, the DiffSpectralNet algorithm proposed in this study shows improved classification accuracy for most ground objects when compared to other classification methods. The proposed method achieves the best OA, AA, and Kappa values, with OA reaching 99.06\%, 99.74\%, and 99.87\% on the IP, PU and SS datasets, respectively. These results indicate that the DiffSpectralNet algorithm efficiently and effectively learns low and high-level features using the diffusion model. Additionally, the DiffSpectralNet algorithm leverages the combination of spectral and spatial information, enabling it to extract a greater amount of information for classification. Therefore, the DiffSpectralNet algorithm proposed in this study demonstrates promising potential for improving the accurate classification of ground objects.

\begin{figure}[!htb] 
    \begin{subfigure}{0.32\textwidth}
    \includegraphics[width=\linewidth]{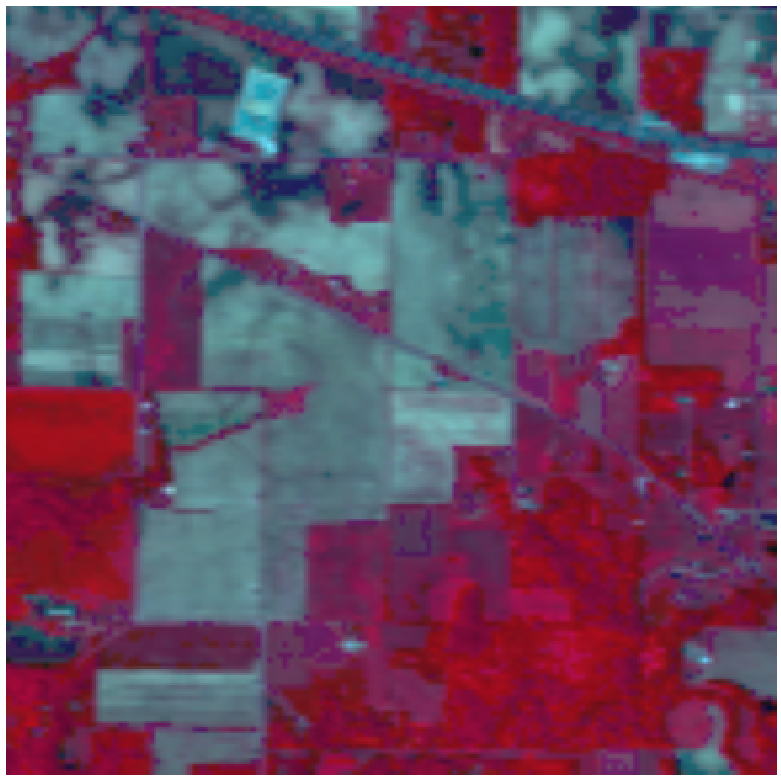}
    \end{subfigure}
    \hfill 
    \begin{subfigure}{0.32\textwidth}
        \includegraphics[width=\linewidth]{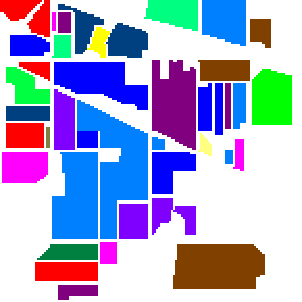}
    \end{subfigure}
    \hfill
    \begin{subfigure}{0.32\textwidth}
        \includegraphics[width=\linewidth]{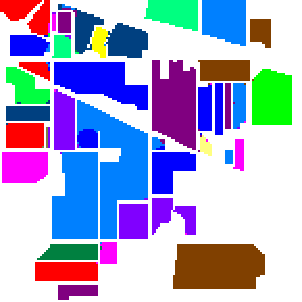}
    \end{subfigure}
    \vspace{1em} 
    \begin{subfigure}{\textwidth}
        \centering
        \includegraphics[width=0.96\linewidth]{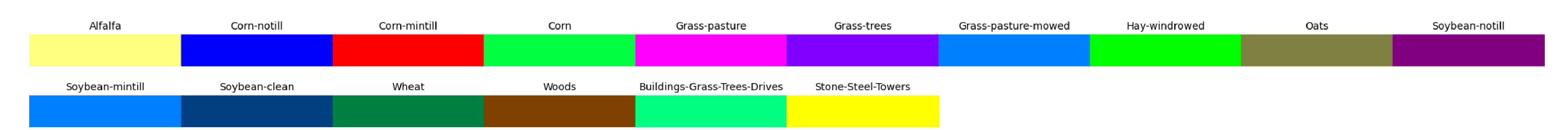}
    \end{subfigure}
    \caption{Classification results of on the IP dataset (a) Original HSI (b) ground truth (c) proposed method } \label{IP_Map}
\end{figure}

\begin{figure}[!htb] 
    \begin{subfigure}{0.32\textwidth}
    \includegraphics[width=\linewidth]{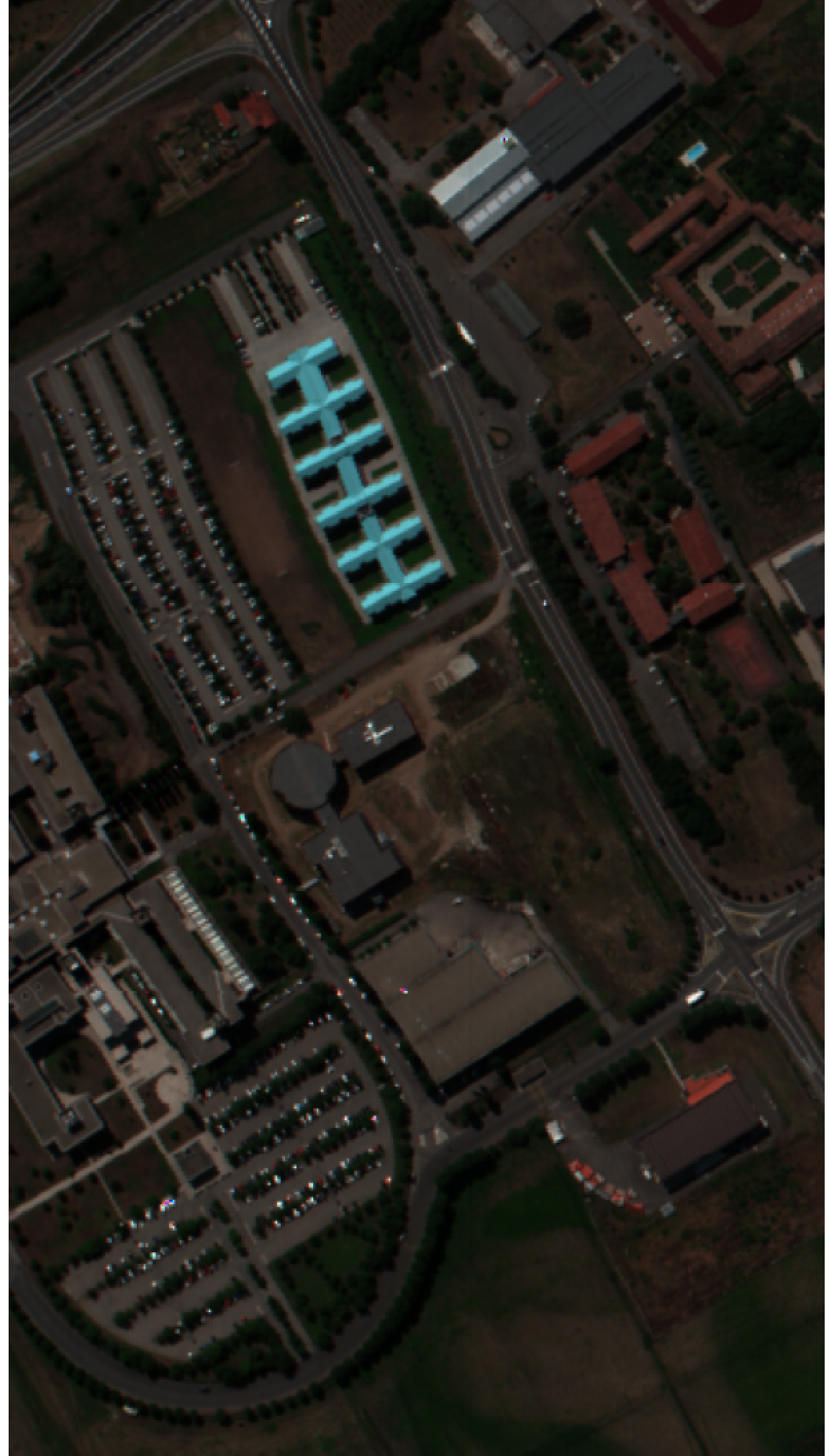}
    \end{subfigure}
    \hfill 
    \begin{subfigure}{0.32\textwidth}
        \includegraphics[width=\linewidth]{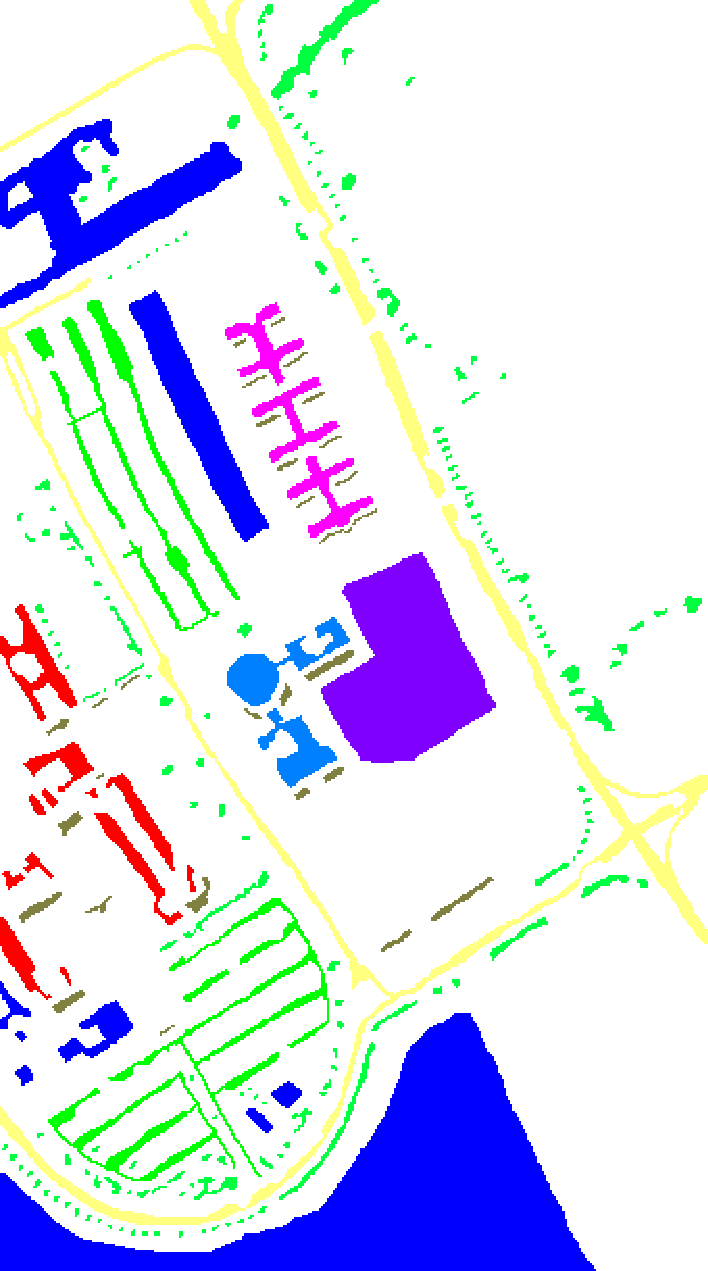}
    \end{subfigure}
    \hfill
    \begin{subfigure}{0.32\textwidth}
        \includegraphics[width=\linewidth]{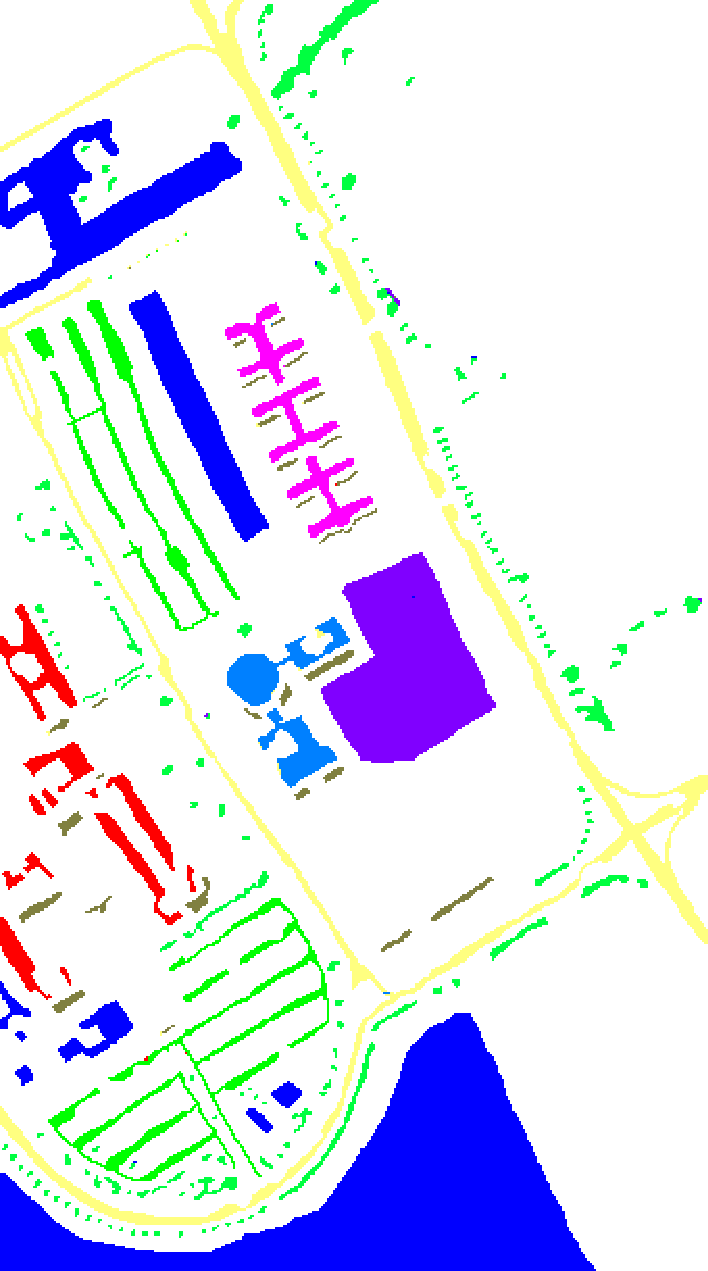}
    \end{subfigure}
    \vspace{1em} 
    \begin{subfigure}{\textwidth}
        \centering
        \includegraphics[width=0.96\linewidth]{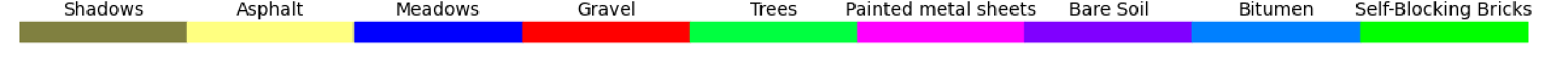}
    \end{subfigure}
    \caption{Classification results of on the PU dataset (a) Original HSI (b) ground truth (c) proposed method } \label{PU_Map}
\end{figure}

\begin{figure}[!htb] 
    \begin{subfigure}{0.32\textwidth}
        \includegraphics[width=\linewidth]{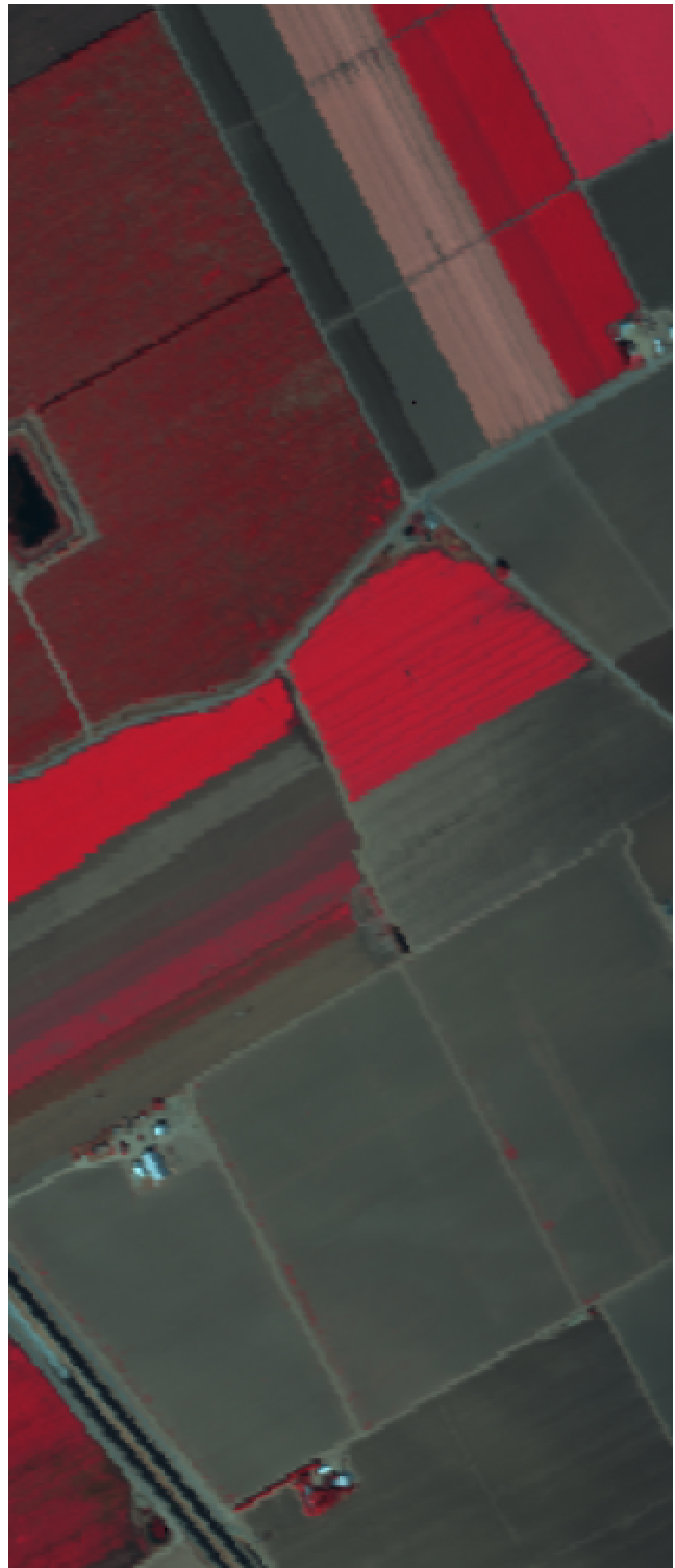}
    \end{subfigure}
    \hfill 
    \begin{subfigure}{0.32\textwidth}
        \includegraphics[width=\linewidth]{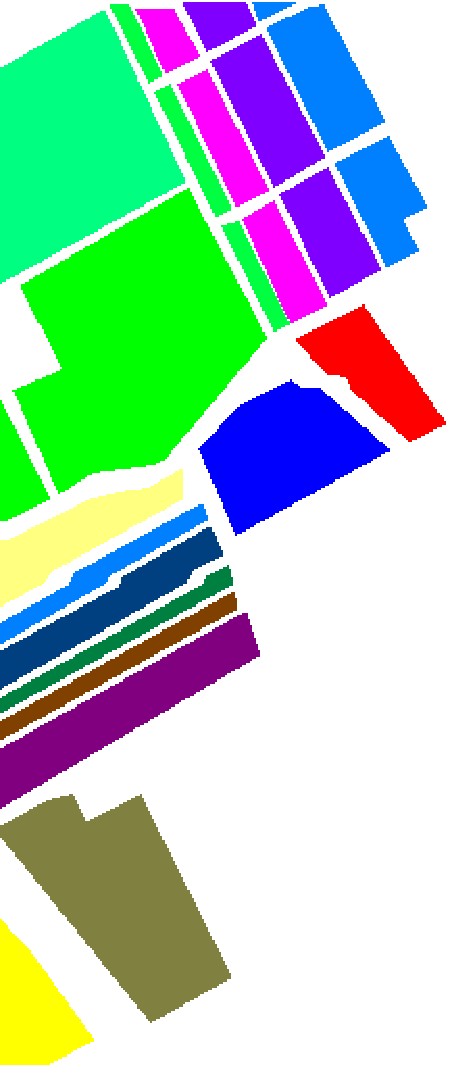}
    \end{subfigure}
    \hfill
    \begin{subfigure}{0.32\textwidth}
        \includegraphics[width=\linewidth]{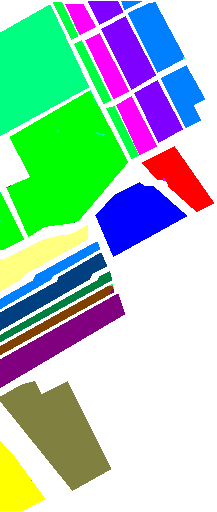}
    \end{subfigure}

    \vspace{1em} 

    \begin{subfigure}{\textwidth}
        \centering
        \includegraphics[width=0.96\linewidth]{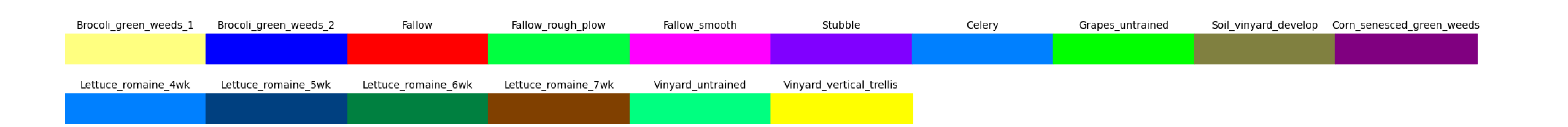}
    \end{subfigure}
    \caption{Classification results of on the SS dataset (a) Original HSI (b) ground truth (c) proposed method} \label{SS_Map}
\end{figure}

In addition to the above quantitative metrics, classification maps in the proposed method have been produced, as shown in Fig. \ref{IP_Map}, Fig. \ref{PU_Map}, and Fig. \ref{SS_Map}. Compared with ground truth, the proposed method obtains more accurate classification results, which further proves the effectiveness of the proposed method in the classification of hyperspectral data.

\textit{Indian Pines Dataset: }
    \begin{table}[h]
    \setlength{\tabcolsep}{3.5pt}
        \caption{Classification accuracies for the proposed and compared HSI classification methods on the IP dataset (the best accuracy in each row is shown in bold).}\label{IP_results}
    \centering
    \begin{tabular}{|c|c|c|c|c|c|c|}
        \hline
       \textbf{Class No.} & \textbf{FuNet-C}  & \textbf{DMVL + SVM}  & \textbf{3DCAE} & \textbf{RPNet--RF}  & \textbf{GSSCRC} & \textbf{Ours} \\

        \hline
        1  & 94.87 & 95.92 & 90.48 & 93.48 & \textbf{100.00} & 87.80 \\
        2  & 68.50 & \textbf{100.00} & 92.49 & 81.30 & 90.97 & 98.67 \\
        3  & 79.59 & \textbf{100.00} & 90.37 & 85.66 & 88.07 & 98.13 \\
        4  & 99.46 & \textbf{100.00} & 86.90 & 83.31 & 84.39 & 97.65 \\
        5  & 95.08 & 94.73 & 94.25 & 94.14 &  95.65 & \textbf{99.54} \\
        6  & 95.70 & 90.96 & 97.07 & 95.15 & 98.77 & \textbf{99.54} \\
        7  & \textbf{100.00} & 98.60 & 91.26 & 43.98 & \textbf{100.00} & \textbf{100.00} \\
        8  & 99.54 & 91.47 & 97.79 & 97.76 & 99.79 &99.77 \\
        9  & \textbf{100.00} & 99.92 & 75.91 & 63.83 & \textbf{100.00} & \textbf{100.00} \\
        10 & 75.93 & 80.63 & 87.34 & 83.07 &  90.53 & \textbf{99.89} \\
        11 & 68.90 & 99.25 & 90.24 & 94.44 & 86.76 & \textbf{99.14} \\
        12 & 71.63 & 94.91 & 95.76 & 82.96 & 94.94 & \textbf{98.69} \\
        13 & 99.38 & 72.49 & 97.49 & 99.10 & 99.51 & \textbf{100.00} \\
        14 & 89.55 & 98.97 & 96.03 & 99.77 & 97.63 & \textbf{100.00} \\
        15 & 91.52 & 95.56 & 90.48 & 98.45 & 79.53 & \textbf{98.85} \\
        16 & \textbf{100.00} & \textbf{100.00} & 98.82 & 97.60 & \textbf{100.00} & 90.48 \\
        \hline
        OA (\%) & 79.89 & 94.60 & 92.35 & 90.23 & 91.33 & \textbf{99.06} \\
        AA (\%) & 89.35 & 94.59 & 92.04 & 87.12 & 93.81 & \textbf{98.00} \\
       \(\kappa\)  & 0.7716 & 0.9400 & - & 0.8887 & 0.9013 & \textbf{0.9893} \\
        \hline
        
    \end{tabular}

\end{table}

    \begin{table}[h]
    \setlength{\tabcolsep}{3.5pt}
        \caption{Classification accuracies for the proposed and compared HSI classification methods on the PU dataset (the best accuracy in each row is shown in bold).}\label{pavia_results}
    \centering
    \begin{tabular}{|c|c|c|c|c|c|c|}
        \hline
{Class No.} & FuNet-C  & DMVL+SVM  & 3DCAE  & RPNet--RF  & GSSCRC & Ours \\

        \hline
        1 & 96.67 & 57.80 & 95.21 & 97.37 &  96.20 & \textbf{99.98} \\
        2  & 97.60 & 98.32 & 96.06 & 99.37 & 98.44 & \textbf{100.00} \\
        3  & 84.49 & 84.37 & 91.32 & 98.19 & 83.66 & \textbf{99.95} \\
        4  & 89.95 & 56.01 & 98.28 & 79.86 &  96.21 & \textbf{98.56} \\
        5 &  99.64 & \textbf{100.00} & 95.55 & 98.85 &  99.63 & \textbf{100.00} \\
        6 & 90.56 & \textbf{100.00} & 95.30 & 99.92 &  93.82 & \textbf{100.00} \\
        7 & 78.27 & \textbf{99.85} & 95.14 & 94.82 & 90.60 & 98.50 \\
        8 & 71.73 & 97.23 & 91.38 & 86.67 & 91.91 & \textbf{99.89} \\
        9 & 98.04 & 27.35 & 99.96 & 99.58 &  \textbf{99.89} & 95.56 \\
        \hline
        OA (\%) & 92.20 & 86.96 & 95.39 & 95.60 & 95.77 & \textbf{99.74} \\
        AA (\%) & 89.66 & 80.10 & 95.36 & 94.96 & 94.13 & \textbf{99.18} \\
       \(\kappa\)  & 0.8951 & 0.8246 & - & 0.9427 & 0.9438 & \textbf{0.9965 } \\
        \hline
        
    \end{tabular}

\end{table}
    \begin{table}[h]
    \setlength{\tabcolsep}{3.5pt}
        \caption{Classification accuracies for the proposed and compared HSI classification methods on the SS dataset (the best accuracy in each row is shown in bold).}\label{sainas_results}
    \centering
    \begin{tabular}{|c|c|c|c|c|c|c|}
        \hline
        \multirow{2}{*}{Class No.} & GSSCRC & DMVL+SVM & 3DCAE & JPPAL\_CRF &  SS1DSwin & Ours \\

        \hline
        1  & \textbf{100.00} & 95.92 & \textbf{100.00} & 99.60  & \textbf{100.00} & \textbf{100.00} \\
        2  & \textbf{100.00} & \textbf{100.00} & 99.29 & 99.87 & 99.97 & \textbf{100.00} \\
        3  & 99.80 & \textbf{100.00} & 97.13  & 98.48 & 99.79 & \textbf{100.00} \\
        4 & 99.92 & \textbf{100.00} & 97.91 & 96.93  & 96.99 & 99.85 \\
        5 & 99.10 & 93.54 & 98.26 & 98.83 &  98.77 & \textbf{99.84} \\
        6 & 99.72 & 99.32 & 99.98 & 99.98 & \textbf{100.00} & 99.89 \\
        7 & 99.86 & 99.39 & 99.64 & 99.92 & 99.94 & 99.79 \\
        8 & 89.76 & 89.94 & 91.58 & 90.16 & 87.67 & 99.74 \\
        9 & \textbf{100.00} & 99.28 & 95.76 & 99.98 &  99.90 & \textbf{100.00} \\
        10 & 98.32 & 94.84 & 96.65 & 94.52 &  95.76 & \textbf{99.94} \\
        11 & 98.65 & \textbf{100.00} & 97.74 & 97.34 & 98.00 & \textbf{100.00} \\
        12 & \textbf{100.00} & 93.82 & 98.84 & \textbf{100.00} & \textbf{100.00} & \textbf{100.00} \\
        13 & 99.13 & 87.23 & 99.26 & 96.11 &  \textbf{100.00} & \textbf{100.00} \\
        14 & 95.70 & 95.98 & 97.49 & 97.87 & 99.00 & 98.13 \\
        15 & 77.43 & 97.50 & 87.85 & 71.32 & 89.27 & \textbf{99.99} \\
        16  & 99.45 & \textbf{100.00} & 98.34 & 98.76 & 99.37 & \textbf{100.00} \\
        \hline
        OA (\%) & 95.62 & 95.88 & 95.81 & 93.34 & 95.45 & \textbf{99.87} \\
        AA (\%) & 97.30 & 96.49 & 97.45 & 96.23 & 97.78 & \textbf{99.82} \\
       \(\kappa\) & 0.9384 & 0.9543 & - & 0.9258 & 0.9493 & \textbf{0.9986} \\
        \hline
        
    \end{tabular}

\end{table}

\subsection{Ablation Studies}

In this section, we analyze the effect of the components in our proposed DiffTrans-HSI.

\begin{enumerate}[label=\arabic*)]
\item \textit{Sensitivity Analysis of Timestep and Feature index: } To analyze the features extracted from the diffusion pre-train model, we have conducted classification experiments on various Timestamp and Featureindex values and recorded the change in the classification performance. When leveraging features from the pre-trained DDPM, two parameters emerge as pivotal: Timestep (T) and FeatureIndex (F). Using the Diffusion model, we monitored classification efficacy alterations as Timestep (t) and FeatureIndex (f) varied. The optimal combination of t and f is essential to ensure accurate outcomes. Table \ref{table 7: sensivity_ip_and_pu} and Table \ref{table 8: sensivity_ss} showcases the performance is sensative to t and f.
For the IP and PU datasets, there is a certain correlation between classification performance and Timestamp or FeatureIndex. When considering the Timestamp dimension, a decreasing trend in classification performance is observed when using features with larger Timestamps, and the optimal performance generally occurs in smaller Timestamp groups. Considering the FeatureIndex dimension, both datasets (IP \& PU) performed better at FeatureIndex 1 than at FeatureIndex 0 and 2.
For SS, there are some fluctuations in classification performance for different Timestamp and FeatureIndex values but no significant changes.
\begin{table}
\centering
\caption{THE PERFORMANCE OF DIFFERENT LAYERINDEX AND TIMESTAMP IN THE INDIAN PINES DATASET, THE PAVIA UNIVERSITY DATASET} \label{table 7: sensivity_ip_and_pu}
\begin{tabular}{cccccccc}
\toprule
FeatureIndex & Timestamp & \multicolumn{3}{c}{Indian Pines} & \multicolumn{3}{c}{Pavia University}  \\
\cmidrule(r){3-5}
\cmidrule(lr){6-8}
& & OA(\%) & AA(\%) & $\kappa $ & OA(\%) & AA(\%) & $\kappa $  \\
\midrule
\multirow{5}{*}{0} & 5 & 98.47 & 95.37 & 0.9826 & 98.94 & 97.93 & 0.9860  \\
& 10 & 98.41 & 96.40 &  0.9818 & 99.15 & 98.68 & 0.9887  \\
& 100 & 97.92 & 96.85 & 0.9762 & 99.03 & 98.27 & 0.9871 \\
& 200 & 97,62 & 94.45 & 0.9728 & 98.63 & 97.91 & 0.9818 \\
& 400 & 98.15 & 96.38 & 0.9789 & 92.86 & 89.98 & 0.9053 \\
\midrule
\multirow{5}{*}{1} & 5 & \textbf{99.06} & \textbf{98.00} & \textbf{0.9893} & \textbf{99.74} & 99.16 & \textbf{0.9965}  \\
& 10 & 98.34 & 96.20  & 0.9811 & 99.63 & 99.09 & 0.9951  \\
& 100 & 98.40 & 96.30 & 0.9817 & 99.54 & \textbf{99.18} & 0.9939  \\
& 200 & 98.45 & 97.48 & 0.9823 & 98.79 & 97.53 & 0.9839 \\
& 400 & 98.29 & 96.35 & 0.9805 & 92.61 & 88.75 & 0.9015 \\
\midrule
\multirow{5}{*}{2} & 5 & 98.59 & 95.17 & 0.9839 & 98.52 & 97.07 & 0.9803 \\
& 10 & 98.82  & 94.99 & 0.9865 & 97.32 & 95.29 & 0.9644 \\
& 100 & 98.01 & 96.05 & 0.9773 & 95.19 & 91.13 & 0.9361  \\
& 200 & 96.37 & 93.26 & 0.9587 & 93.54 & 90.25 & 0.9139  \\
& 400 & 95.71 & 92.52 & 0.9510 & 86.66 & 81.28 & 0.8202 \\
\bottomrule
\end{tabular}
\end{table}

\begin{table}
\centering
\caption{THE PERFORMANCE OF DIFFERENT LAYERINDEX AND TIMESTAMP IN THE SALINAS SCENE DATASET}\label{table 8: sensivity_ss}
\begin{tabular}{cccccccc} 
\toprule
FeatureIndex & Timestamp & \multicolumn{3}{c}{Salinas Scene}   \\
\cmidrule(r){3-5}
\cmidrule(lr){6-8}
& & OA(\%) & AA(\%) & $\kappa $  \\
\midrule
\multirow{5}{*}{0} & 5 & 99.74 & 99.73 & 0.9971  \\
& 10 & \textbf{99.87} & \textbf{99.82} &  0.9985  \\
& 100 & 99.71 & 99.67 & 0.9967 \\
& 200 & 98.63 & 97.91 & 0.9818  \\
& 400 & 98.29 & 97.74 & 0.9809 \\
\midrule
\multirow{5}{*}{1} & 5 & 99.83 & 99.76 & 0.9981  \\
& 10 & 99.76 & 99.73  & 0.9973   \\
& 100 & \textbf{99.87} & 99.81 & \textbf{0.9986}  \\
& 200 & 98.45 & 97.48 & 0.9823 \\
& 400 & 98.06 & 97.70 & 0.9784  \\
\midrule
\multirow{5}{*}{2} & 5 & 99.26 & 99.32 & 0.9917 \\
& 10 & 98.95  & 99.00 & 0.9883  \\
& 100 & 98.04 & 97.98 & 0.9782  \\
& 200 & 96.37 & 93.26 & 0.9587  \\
& 400 & 91.84 & 88.39 & 0.9089  \\
\bottomrule
\end{tabular}
\end{table}

\item \textit{Percentage of Training Samples: } 

It is widely known that the number of training samples directly affects the performance of the network. To verify this with the proposed DiffSpectralNet, we randomly evaluated different proportions (5\%, 10\%, 15\% and 20\%) of the entire training dataset. and depict the comparative results in \textcolor{red}{Fig. }. As expected, the classification accuracy gradually improves with an increase in training samples. It is worth noting that OA tends to be stable when the percentage of training samples is greater than 80\%. However, when the percentage of training samples in the Indian Pines dataset is less than 40\%, the performance is unsatisfactory due to the insufficient number of samples for proper training. Therefore, it is reasonable to extrapolate that DiffSpectralNet is reliable and stable for this task.

\end{enumerate}

\section{Conclusion}

In this research, we present the DiffSpectralNet technique for classifying hyperspectral images. Hyperspectral images (HSI) contain a wealth of spectral-spatial information and complex relationships between bands. Most existing methods for HSI classification rely on CNN or Transformer models, but they may not efficiently extract patterns and information. Our proposed method effectively and efficiently learns discriminative spectral-spatial features using the diffusion model. This approach allows us to explore and utilize the spatial-spectral neighborhood structure of hyperspectral data, resulting in the effective extraction of deep features.

To gather spectral information, we employ a transformer-based model with a cross-layer skip connection. We demonstrate the superiority of our proposed Diff-HSI approach by achieving state-of-the-art results in HSI classification based on quantitative trials conducted on three HSI datasets.

However, it's important to note that while our classification techniques show promise in hyperspectral image classification, we have not yet investigated their generalizability beyond this specific context. In future studies, we plan to further validate and enhance the performance of our proposed model on additional hyperspectral datasets.




\bibliography{sn-bibliography}

\end{document}